%% file: paper.tex
\documentclass[letterpaper, 10 pt, conference]{ieeeconf}  

\IEEEoverridecommandlockouts                              

\usepackage{graphics} 
\usepackage{epsfig} 
\usepackage{mathptmx} 
\usepackage{times} 
\usepackage{amsmath} 
\usepackage{amssymb}  
\usepackage[caption=false,font=footnotesize]{subfig}
\usepackage{color}

\title{\LARGE \bf
Can Modular Finger Control for In-Hand Object Stabilization be accomplished by Independent Tactile Feedback Control Laws? }

\author{Filipe Veiga$^{1}$ and Jan Peters$^{1,2}$
\thanks{$^{1}$Filipe Veiga and Jan Peters are with the Technische Universit\"{a}t Darmstadt, Darmstadt, Germany, FG Intelligent Autonomous Systems. e-mail:  \{veiga, peters\}@ias.tu-darmstadt.de}%
\thanks{$^{2}$Jan Peters is also with the Max-Planck-Institut f\"{u}r Intelligente Systeme, T\"{u}bingen, Germany.}
}

\begin{document}

\maketitle
\thispagestyle{empty}
\pagestyle{empty}

\input{abstract}

\input{Introduction}

\input{PredictingSlip}

\input{ExperimentalEvaluation}

\input{DiscussionConclusion}

\input{Acknowledgements}

\bibliographystyle{IEEEtran}
\bibliography{bibliography}

\end{document}

%% file: abstract.tex
\begin{abstract}
Currently grip control during in-hand manipulation is usually 
modeled as part of a monolithic task, yielding complex controllers
based on force control specialized for their situations. 
%
Such non-modular and specialized control approaches render the 
generalization of these controllers to new in-hand manipulation tasks
difficult. Clearly, a grip control approach that generalizes well between 
several tasks would be preferable.
%
We propose a modular approach where each finger is controlled by an independent
tactile grip controller. Using signals from the human-inspired biotac sensor, we can 
predict future slip -- and prevent it by appropriate motor actions. This slip-preventing
grip controller is first developed and trained during a single-finger stabilization task.
Subsequently, we show that several independent slip-preventing grip controllers
can be employed together without any form of central communication. The resulting
approach works for two, three, four and five finger grip stabilization control. 
%
Such a modular grip control approach has the potential to generalize across
a large variety of in-hand manipulation tasks, including grip change, finger gaiting,
between-hands object transfer, and across multiple objects. 
\end{abstract}

%% file: Introduction.tex
\section{Introduction}
\label{sec:intro}

%
%
%

To date, most grasping and in-hand manipulation approaches are based on global 
monolithic planning and control strategies. These monolithic policies determine
finger placement, grasping configuration, finger trajectories, contact forces and 
contact locations for the entire hand throughout the entire manipulation task \cite{hertkorn2013planning,van2015learning}.
Such an approach requires highly accurate kinematic/dynamic models of both hand and object along 
with precise sensing of positions (of both hand and object) as well as interaction
forces. In practice, uncertainty is frequently too high for jointly using all of these 
components and, thus, the focus of grasping has become largely data-driven \cite{Bohg2013}.

Unfortunately, data-driven approaches do not come for free but either require a large training data set,
retricting the tasks to sufficiently similar scenarios \cite{Bohg2013} or a low-dimensional representation
that covers the considered manipulation tasks (such as synergies~\cite{prattichizzo2010motion},
motion primitives \cite{kazemi2012robust}, etc.). All data-driven solutions are obviously task- and platform-specific.
The learned polices inherently couple the employed degrees of freedom, rendering the 
incorporation of tactile feedback from multiple fingers difficult.

To enable a control policy to both deal with uncertainty (e.g., in contact locations and forces) and
to generalize well beyond a limited set of cases, an approach needs to be both data-driven and 
modular. Human grasping and manipulation appears to largely data-driven~\cite{Johansson2009}
but there is strong evidence that the resulting control strategies are just based on local 
sensing~\cite{edin1992independent}, rendering the control of the fingers largely independent 
from each other~\cite{edin1992independent} thus, being highly modular. The correlations 
in human movement appear to arise from higher-level planning instead of low-level control~\cite{burstedt1997coordination}. 
Clearly such an approach would be desirable in robotic grasping and manipulation. 
 
\begin{figure}[b!]
    \centering
    \subfloat[Wessling Robotics Hand]{\label{fig:random_angles} \includegraphics[width=0.49\linewidth]{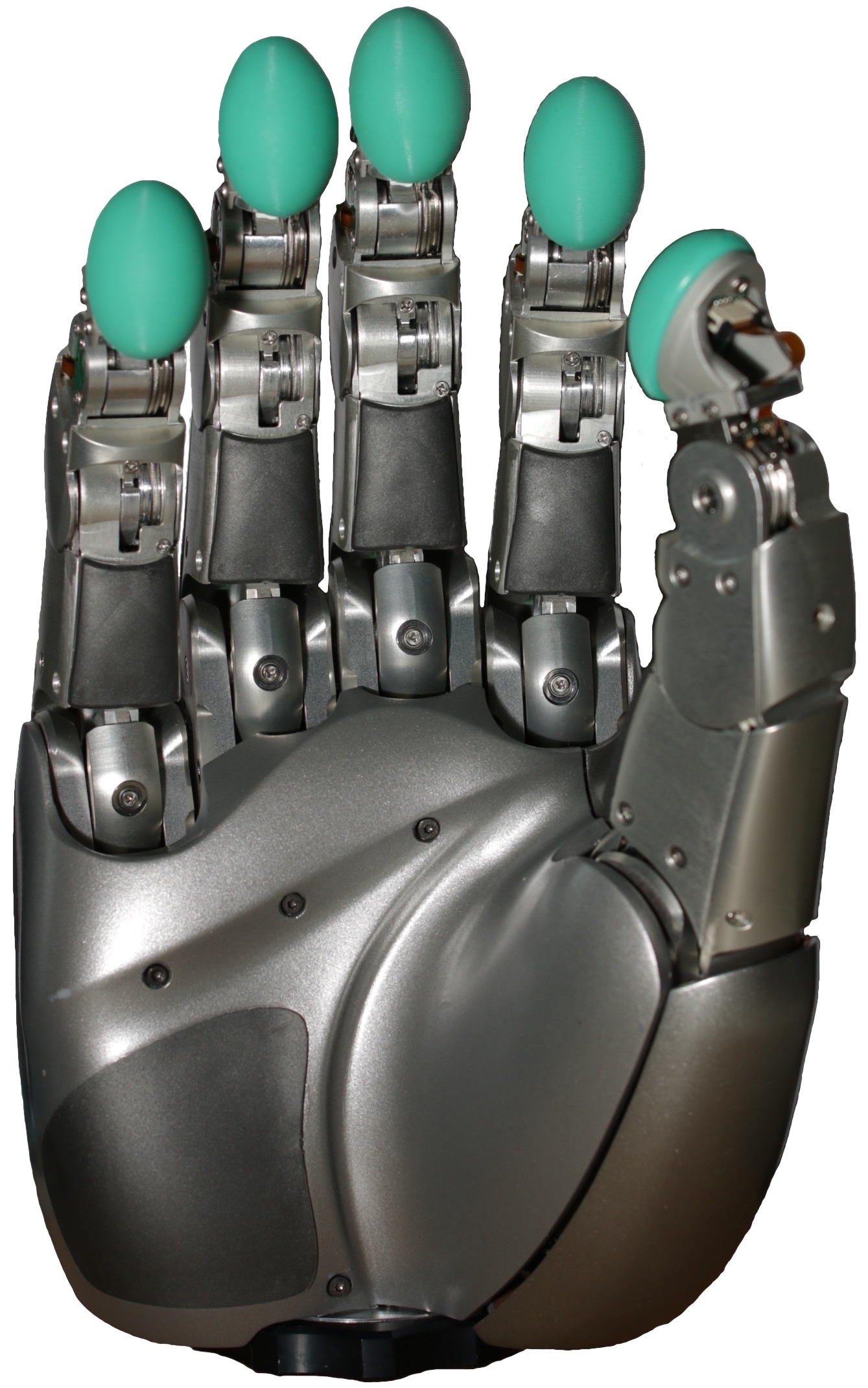}}
    \subfloat[Allegro Hand]{\label{fig:random_axis} \includegraphics[width=0.49\linewidth]{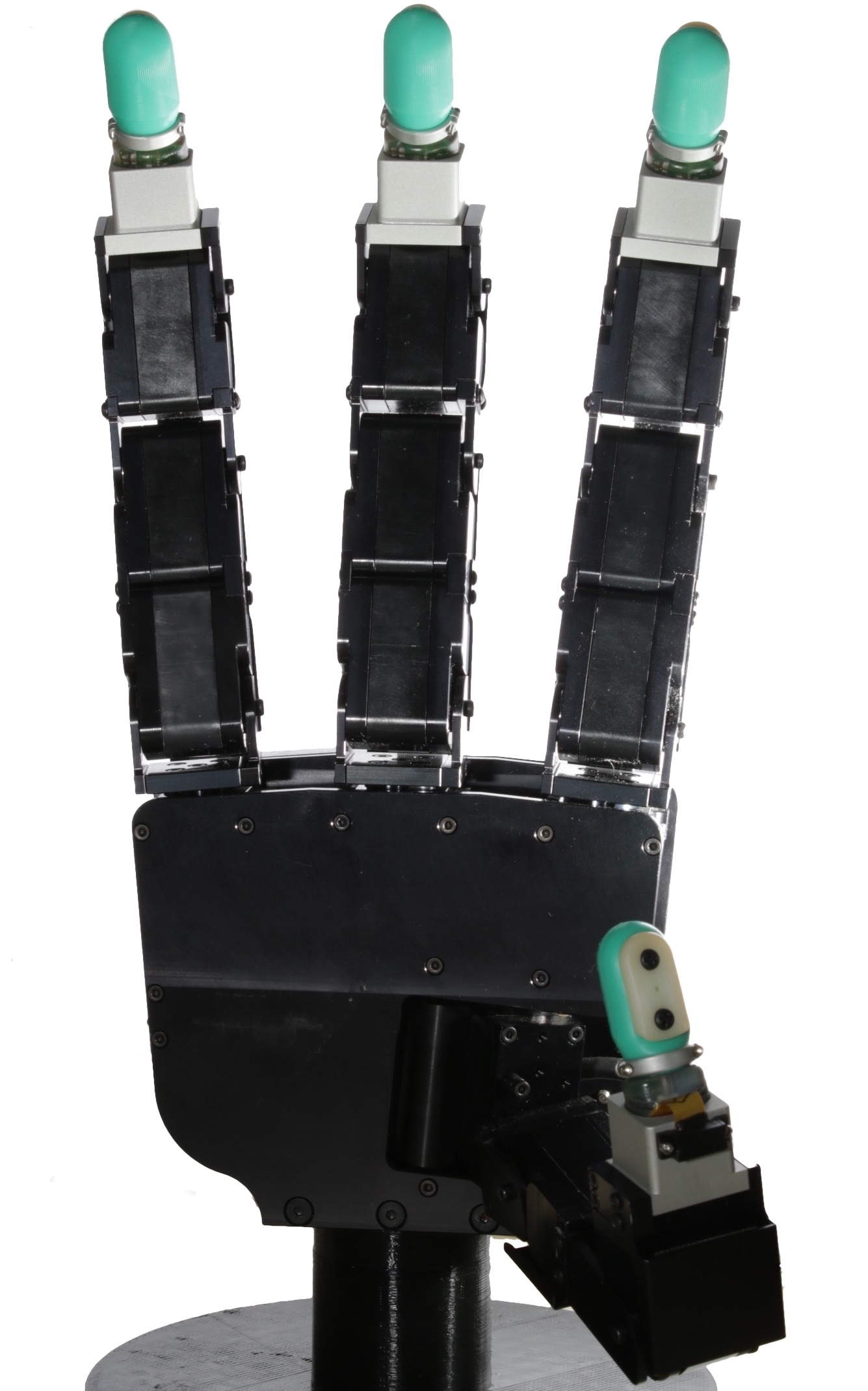}}
    \caption{\label{fig:our:hands}
    	The controllers proposed in this paper were successfully evaluated both on the Wessling Robotics Five Finger Hand and the Allegro Four Finger Hand. 
    	Each hand only has identical fingers, which allows the learned models to be transferred easily across the four or five 
    	fingers of the respective hands. Different versions of the BioTac sensor are employed in both hands, i.e., the regular BioTac
    	with 19~electrodes and the BioTac SP with 22~electrodes.}
\end{figure}

Ensuring grip stability of the object in the hand is frequently central to both grip control (i.e., stabilizing an object)
and in-hand object manipulation (i.e., moving an object between stable grip configurations.
Classical robotics approaches often rely on measures such as form- or force-closure for assessing grip stability;
clearly for imperfect models and imperfect contact/force sensing, the use of such measures is not straightforward. 
Many researchers have proposed alternative grasp stability measures~\cite{bekiroglu2011assessing, dang-auro2013, Madry2014, li-iros2014-adaptation, Romano2011}
and developed accompanying control strategies.
For unknown objects, it may make more sense to reactively prevent instability than to maximize a potentially arbitrary stability measure~\cite{veigastabilizing}.  
In grasping and in-hand manipulation, instability is synonymous with slip~\cite{tremblay-icra1993}.  While slip cannot be directly measured, it can be
detected and predicted based on high-frequency tactile sensing~\cite{veigastabilizing}. We propose that a grip stabilizing control law 
should aim to ensure that there will be no future slip.  


Inspired by progression from one-finger over two-fingers to the whole hand
proposed by~\cite{Klatzky1987object} in the context of tactile object exploration and
by the independent control hypothesis in human grip control by~\cite{edin1992independent,burstedt1997coordination},
we aim to develop independent control policies based on tactile feedback for each finger 
that in conjunction generalize from one-finger to five-finger gripping and easy in-hand 
manipulation.\vspace{0.2cm}
\vspace{0.2cm}\framebox{\parbox[t][2.0cm]{0.98\linewidth}{\centering \bf \textsl{Edin/Burstedt's Hypothesis~\cite{edin1992independent,burstedt1997coordination}}\\
While human grasp planning is global, human grib control is local in every finger -- they ``communicate'' through the object. It does not really matter whether the fingers belong to different humans.
}}
To accomplish this goal, we endow every finger with learned predictive models of 
future slip based on the high-frequency features of the tactile finger-object interaction. 
The local control law in the finger chooses the finger's movement such that probability 
of future slip is minimized. The resulting control law is capable stabilizing objects against 
other objects (such as a table or a wall), jointly stabilizing objects with more robotic 
fingers (as in in-hand object stabilization or gripping) or against the hand of a human 
operator (human-robot joint stabilization). It can also be employed for in-hand manipulation 
by stabilizing an object with several fingers while one or more fingers move the object 
within the stable grip. The coordination between modular finger controller occurs only 
indirectly through the tactile signals observed by each finger. 

This modular approach enables a higher-level planning system to operate 
with less object knowledge while requiring fewer complex models for control  
than analytical approaches. The proposed reactive control framework
generalizes potentially better than monolithic data-driven approaches 
across multiple tasks, a variety of objects and different robotic platforms.
Our approach reproduces findings in human motor control where the 
absolute amount of force applied by single digits will always settle just 
above the minimal amount of forces to prevent slip~\cite{Johansson2009,edin1992independent}
in static setting.\vspace{0.2cm}
\framebox{\parbox[t][2.5cm]{0.98\linewidth}{ \centering \bf \textsl{Our Working Hypothesis} \\
If a single finger controller can enable a human and a robot finger to hold objects well-together based only on actile sensing, then a modular $N$-finger grip control can be accomplished through $N$ independent tactile sensing-based finger grip control.}}

%% file: PredictingSlip.tex
\section{Modular Tactile Sensing-based \\ In-Hand Object Stabilization} 
Classical approaches to sensing-based in-hand stabilization of objects have relied on accurate force-torque sensors, precise models of the held objects, optimal contact points, highly accurate finger control and the central computation of joint-level forces to be applied in order to accomplish stable force closure. Such requirements clearly cannot always be met and simpler approaches to sensing-based in-hand stabilization are desirable. In this paper, we propose a novel approach that goes into the opposite direction: We assume that single fingers can accomplish stabilization without central control purely relying on the prediction of slip based on tactile sensing and, thus, can be considered independent modules. As a first step, we introduce a single finger tactile control approach which works well for stabilizing objects pinned against other objects. Subsequently, in a second step, we discuss how this approach can be used in conjunction with more finger -- from its own hand as well as from another agent; see Fig.~\ref{fig:single:finger:stabilization} for an experiment where this method stabilized an object jointly with a human finger. 

\subsection{Single-Finger Stabilization based on Slip Prediction \label{sec:single:finger:stabilization}}
Humans appear to have two components during tactile stabilization of objects, i.e., they assess the potential danger of slip 
and reactively adjust the applied forces until slip is no longer likely \cite{edin1992independent}. Thus, to obtain our
single-finger tactile sensing-based object stabilization controller, we require two components, i.e., (1)~a slip predictor
and (2)~a force adjustment method based on slip prediction.  

\subsubsection{Slip Prediction} We can formalize such an 
approach by treating slip prediction as a classification problem. Here, a classifier \(f(\cdot)\) is employed to 
predict the state at time $t+\tau_{f}$ as slip or not slip (a preliminary version of this idea has been presented in~\cite{veigastabilizing}). 
To predict slip in such a way, features $\phi(\cdot)$ of the raw sensor values \(\mathbf{x}_{(t-\tau_H):t}\) need to be extracted for a
window in time (i.e., the last $\tau_H$ time steps). Thus, we need to train the slip predictors
\begin{equation}
c_{t+\tau_{f}} = f(\phi(\mathbf{x}_{(t-\tau_H):t}))
\label{eq:prediction model}
\end{equation}
using $c_{t+\tau_{f}} \in \{\texttt{slip}, \texttt{contact}, \neg \texttt{contact} \}$ as labels for the prediction at time-step $t+\tau_{f}$ where $\tau_{f}$ is the prediction horizon, i.e.,  it denotes the number steps into the future where slip is predicted.

\begin{figure}[t]
    \centering
    \includegraphics[width=\linewidth]{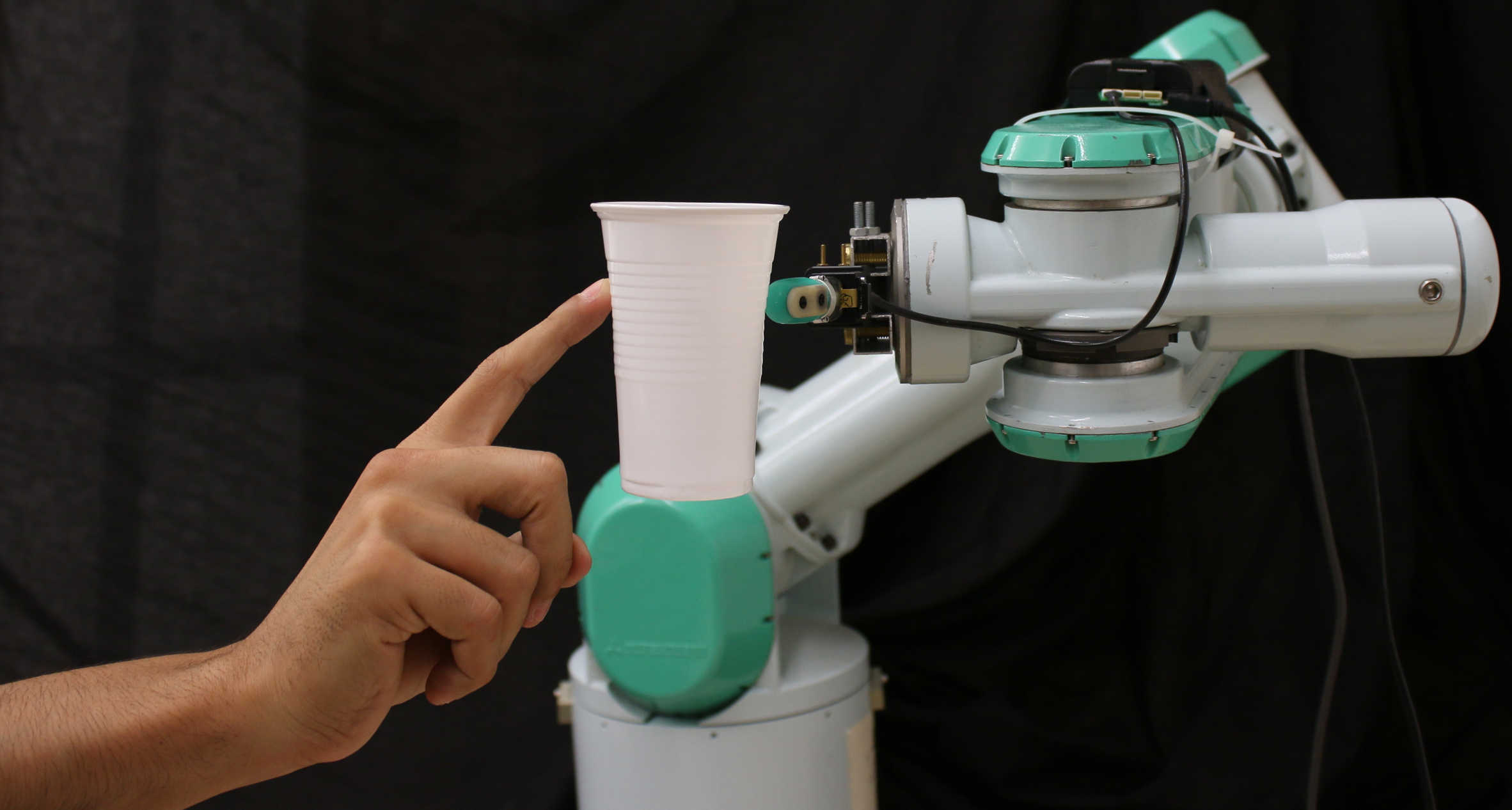}
    \caption{\label{fig:single:finger:stabilization} 
    	Single finger stabilization (as described in Sec.~\ref{sec:single:finger:stabilization}) can be decomposed into a slip prediction 
	component and a control law that attempts to adjust all commands to obtain the minimum force where no slip occurs. 
	Even when the slip predictors are trained by pinning objects against a fixed plane, these slip predictors generalize to 
	multi-finger scenarios. Here, we depict a PA10 used as a single that predicts slip based on a single BioTac sensor
	but successfully stabilizes an object together with a human finger. The same predictors and controller are used in the 
	multi-finger scenarios through-out this paper.}
\end{figure}

\subsubsection{Force adjustment} \label{sec:force-adjustment}
We require a control law that converts the predicted slip label $c_{t+\tau_{f}}$ into joint velocities
$\mathbf{\dot{q}}_{t}$, as we only have velocity control access for both our robot hands (i.e., the Allegro hand and the Wessling five finger hand in Fig.~\ref{fig:our:hands})
and for the PA-10 robot arm (in Fig.~\ref{fig:single:finger:stabilization}). Inspired by neuroscientific studies~\cite{edin1992independent}, 
our controller only maintains a sufficient statistic on the past slippage of the finger, i.e., 
\begin{equation}
	l_{t+1} = \begin{cases} 
		l_{t} + s_{\textrm{slip}} 		& \mbox{if}\; c_{t+\tau_{f}} = \texttt{slip}, \\
		l_{t} - s_{\neg \textrm{slip}} 	&  \mbox{if}\; c_{t+\tau_{f}} = \texttt{contact},\\
		l_{t}  					 	&  \mbox{if}\; c_{t+\tau_{f}} = \neg\texttt{contact},
	\end{cases}
\end{equation} 
which basically is the weighted balance between the time where the object slip is predicted and where not. Here, adding a weight $s_{\textrm{slip}}$
should result in the controller to apply more force to the object while subtracting $s_{\neg \textrm{slip}}$ will reduce the amount of applied forces.   
We employ the resulting statistic $l_{t}$ in an exponential regulator that increases the desired velocity of the finger tip towards 
the contact direction if slip is frequently predicted, implicitly increasing the the normal contact force. The velocity quickly goes to jittering around 
zero when no slip is predicted. These requirements result into the control law  
\begin{equation}
\mathbf{\dot{q}}_{t} = \beta\mathbf{N}(\mathbf{x}_{t})e^{\alpha l_{t}}
\end{equation}
where $\beta>0$ and $\alpha>0$ are controller parameters and $\mathbf{N}(\mathbf{x}_{t})$ denotes the estimated contact normal at time $t$.

\subsection{Multi-Finger Gripping by Single-Finger Predicted Slip Control}
\label{sec:multi-finger-scenario}
When progressing towards in-hand stabilization and in-hand manipulation, more fingers are required and the complexity of the tasks quickly scales accordingly with the hand dexterity. Generally, a higher dimensionality can be coped with either by identifying a lower-dimensional manifold for the problem or by decomposing the problem. Following the core insight in~\cite{edin1992independent} that human multi-finger grip stabilization appears to be accomplished by separate neural circuits that interact through the object instead of via the central nervous system. Following their insight, we hypothesize that \textsl{multi-finger robot gripping can be accomplished using the same single-finger predicted slip controller on all fingers independently from each other}. As a first scenario, we reproduce the scenario in~\cite{edin1992independent}, where two humans jointly hold an object using one finger each -- with the same apparent ease as if one human was using one finger from each hand. The underlying neural control appears to be unaffected, no matter whether it interacts with finger from the same human or another human. We reproduce this effect replacing one human finger by a robot finger in a human-robot joint stabilization task as shown in Fig.~\ref{fig:single:finger:stabilization} -- and the single finger controllers of both the human and the natural finger directly work well together. 

Thus, to fully utilize the dexterous capabilities of the hand, we propose that each hand should be seen as a set of independently controlled fingers\footnote{Note that this does not imply independent planning, no independent making and breaking contact but only the stabilization phase.}. 
A set of independent fingers, in contrast to a fully connected manipulator, allows decomposing the object stabilization control problem such that each finger separately predicts future slip based on tactile sensing and avoids it by independently adjusting the applied forces. While it may appear that synchronization only through the tactile feedback may appear counter intuitive, it actually not only greatly reduces the problem dimensionality but also ensures that the fingers only influence each other when it is needed for the stabilization of the object. As a result, it not only becomes more straightforward to design stabilizing control laws but the synchronization becomes more robust. 

%

%% file: ExperimentalEvaluation.tex
\section{Experimental Evaluation}
\label{sec:exp-eval}

To explore the capabilities of the concept of finger independence discussed in Sec.\ref{sec:multi-finger-scenario} when using the control strategy proposed in Sec.\ref{sec:force-adjustment}, we perform several experiments. We attempt to stabilize the object with a varying number of fingers with both the Allegro and the Wessling hands, Sec.~\ref{sec:grip-stabilization-evaluation}. The showcase of aforementioned experiments is preceded by a detailed description of the two robotic platforms used, Sec.\ref{sec:testing-platforms}, an account of the tactile sensors mounted on these platforms and their differences, Sec.\ref{sec:tactile-sensors}, and a detailed outline of the procedure used to acquire the ground truth data for the slip classifiers, Sec.~\ref{sec:tactile-training-data}. 

\subsection{Experimental Setup: Testing Platforms \& Tactile Sensors}

We use two different robotic hands to assess the generalization capabilities of our approach across different robots. Each hand as sensorized fingertips, equipped with two different versions of the BioTac sensors. These two versions of the sensors, although similar, have significant differences with respect to shape and sampling frequencies. The Allegro and Wessling hands are introduced in Sec.\ref{sec:testing-platforms} and the two versions of the BioTac sensors are described in Sec.\ref{sec:tactile-sensors}.

\subsubsection{Testing Platforms}\label{sec:testing-platforms}

Two hands are used to test our stabilization controllers as we which to show that the proposed approach is platform independent. Both hands are depicted in Fig.~\ref{fig:our:hands}

The \textsl{Allegro Hand}, produced by Wonik Robotics, is a lightweight four fingered hand with four joints per finger, for a total of 16 actuated degrees of freedom. The thumb has an abduction joint, two metacarpal joints (rotation and flexing) and a proximal joint. The remaining fingers do not have metacarpal rotation joints and instead have a distal joint. Each finger is equipped with a BioTac sensor on the fingertip.

The \textsl{Wessling Robotic Hand}, produced by Wessling Robotics, is composed of five modular designed robotic fingers. Each finger consists of three actuated degrees of freedom and coupled joint, offering 15 actuated degrees of freedom and 20 joints in total. The hand's fingertips are equipped with BioTac SP sensors.
 
\subsubsection{Tactile Sensors}\label{sec:tactile-sensors}

\begin{figure}[t]
    \centering
    \subfloat[BioTac]{\includegraphics[width=0.49\linewidth]{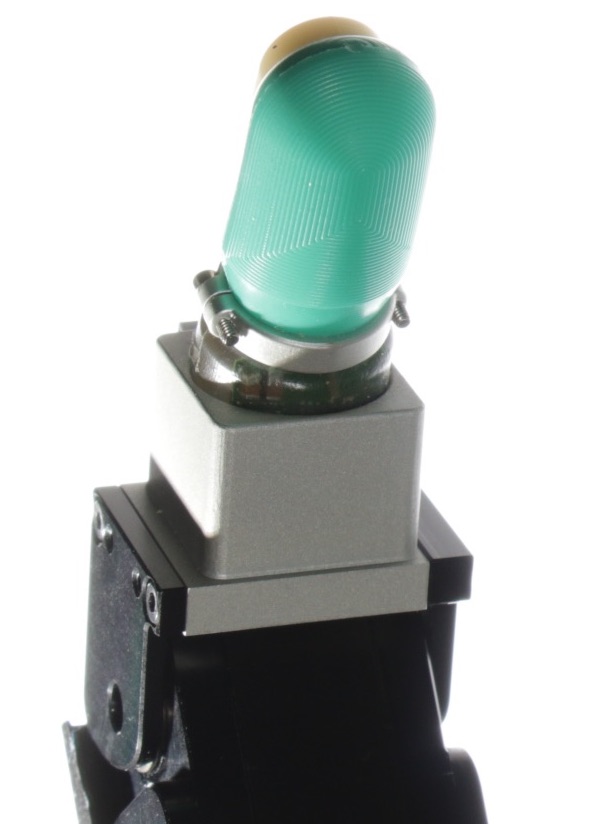}}
    \subfloat[BioTac SP]{\includegraphics[width=0.49\linewidth]{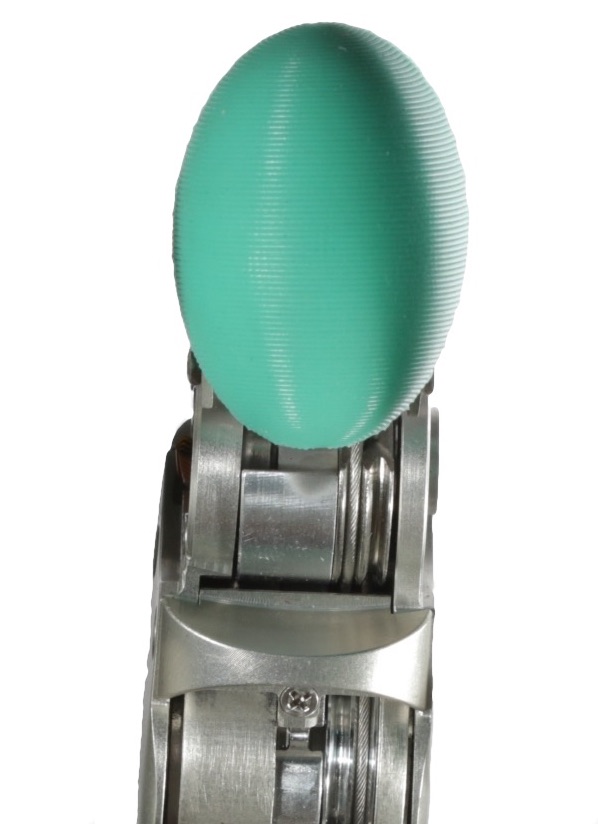}}
    \caption{\label{fig:our:sensors}Syntouch's BioTac and BioTac SP fingertip sensors respectively equipping the Allegro and Wessling hands fingers. These multi-model sensors i. provide information on finger pressure using pressure transducers, ii. temperature and temperature flow via a set of heaters in conjunction with a thermistor, and iii. local deformation through impedance-sensing electrodes distributed on the fingers surface.}
\end{figure}

The raw tactile data is extracted from the fingertip sensors equipped on each hand. The BioTac and BioTac SP, respectively for the Allegro and Wessling Hands, produced by Syntouch, LLC. Both sensors are multi-channel tactile sensors inspired by the human finger and provide multi-modal responses composed of low frequency $P_{\textrm{dc}}$ and high frequency $P_{\textrm{ac}}$ pressure, temperature $T_{\textrm{dc}}$ and temperature flow $T_{\textrm{ac}}$ and local skin deformation $E$. The sensors can be seen on the respective fingertips in Fig.~\ref{fig:our:sensors}

The BioTac, acquires its pressure signals $P_{\textrm{dc}}$ and $P_{\textrm{ac}}$ using a pressure transducer. Nineteen impedance-sensing electrodes $E \in \mathbb{R}^{19}$ are spread across the finger surface measuring local deformation. Finally, $T_{\textrm{dc}}$ and $T_{\textrm{ac}}$ are measured using a set of heaters coupled with a thermistor. All channels are sampled at a rate of 100~Hz. The $P_{\textrm{ac}}$ is acquired by the sensor at a rate of 2.2~kHz, but is still sampled at 100~Hz, producing batches of \(22\) values every 10~ms. 

In comparison, the BioTac SP differs from the BioTac in two aspects. Firstly, the sensor is designed with an elliptical shape influencing the number of electrodes that can be distributed on the finger surface. The SP's shape allows twenty two electrodes $E_{SP} \in \mathbb{R}^{22}$ to be distributed on it's surface, in contrast to the nineteen in the BioTac. The SP then as a denser sensing of local finger deformation. The other difference is the sampling frequency of the sensor. For the SP, the high frequency pressure, $P_{\textrm{ac}}$, is sampled at 4.545KHz, the electrodes $E_{SP}$ and the pressure $P_{\textrm{dc}}$ are sampled at 1KHz and the temperatures $T_{\textrm{ac}}$ and $T_{\textrm{dc}}$ are sampled at 200Hz. This allows for a much faster feedback loop on than the one available when use the BioTac.

\subsection{Tactile Training Data}
\label{sec:tactile-training-data}
For training the slip classifiers, data has to be collected on both platforms as they are equipped with similar yet different sensors. The data collection procedure is identical for both platforms.

\begin{figure}[t]
    \centering
    \includegraphics[width=\linewidth]{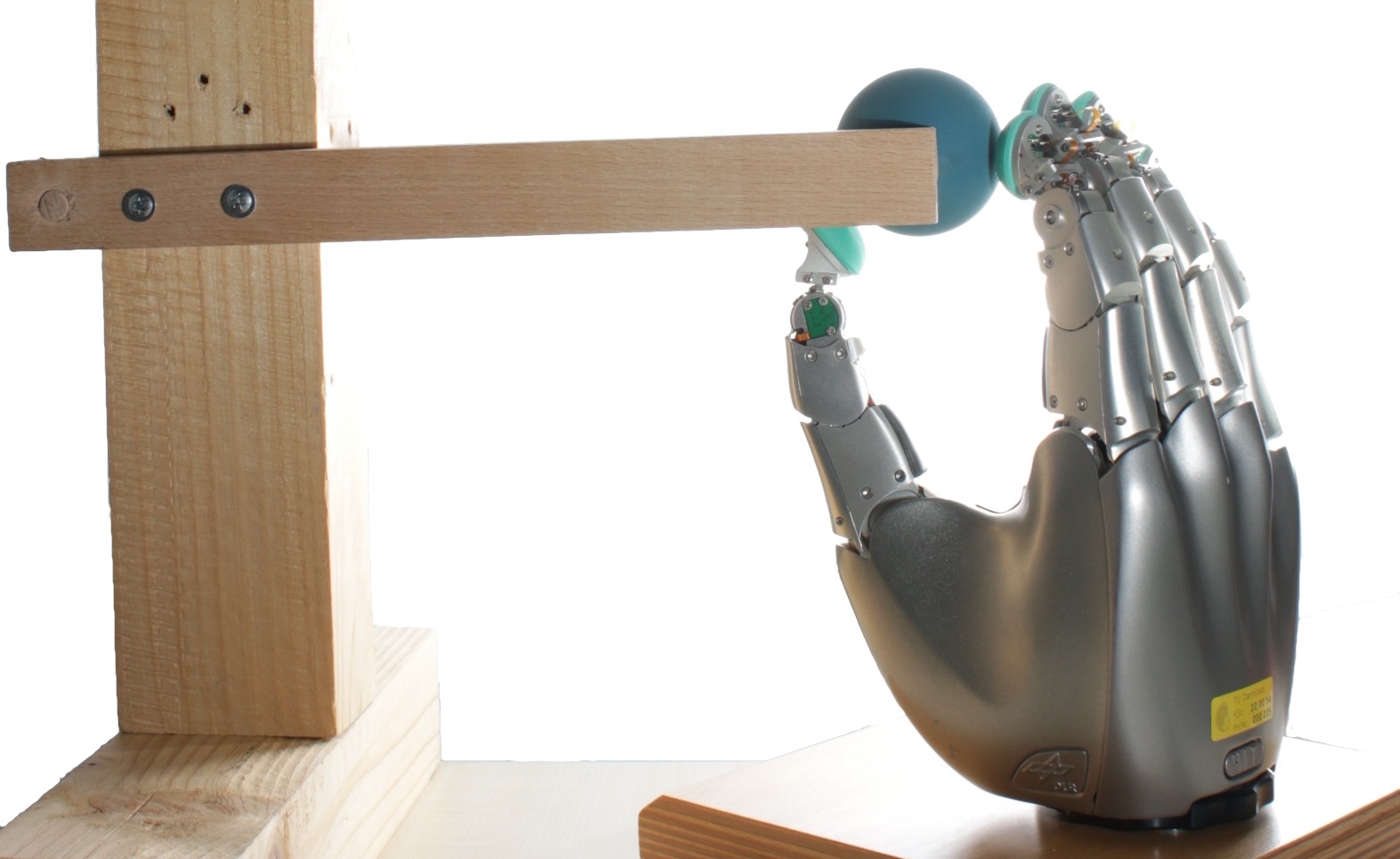}
    \caption{\label{fig:data:collection}The data collection setup used for acquiring the necessary training data for the slip classifiers. In this particular example, the Wessling Hand was used. An object is fixed by two wooden plates while in the hand's work space. After being initialized in a predefined position, three fingers are flexed until each becomes in contact with the object, having their pressure immediately adjusted to a target pressure. In a final step, the three finger survey the object's surface with random direction and velocity.}
\end{figure}

Firstly an object is fixed to a support that holds it on the hand's work space. All fingers are positioned in an initial configuration and are subsequently flexed until each becomes in contact with the object. After the fingers are in contact with the object, the pressure applied by each finger is adjusted by a PID controller until a target pressure is reached on all fingers. Finally, small surveying motions are performed on the object surface by three fingers simultaneously. Acquiring data from three sensors simultaneously reduces the necessary number of training trials. All data from each of the fingers is concatenated into a single data set that will be used to train each of the individual slip predictors. The data collection setup is exemplified in Fig.~\ref{fig:data:collection} for the Wessling hand data acquisition. 

\begin{figure*}[p]
    \centering
    \vspace{-1cm}
    \subfloat[]{\includegraphics[width=0.15\linewidth]{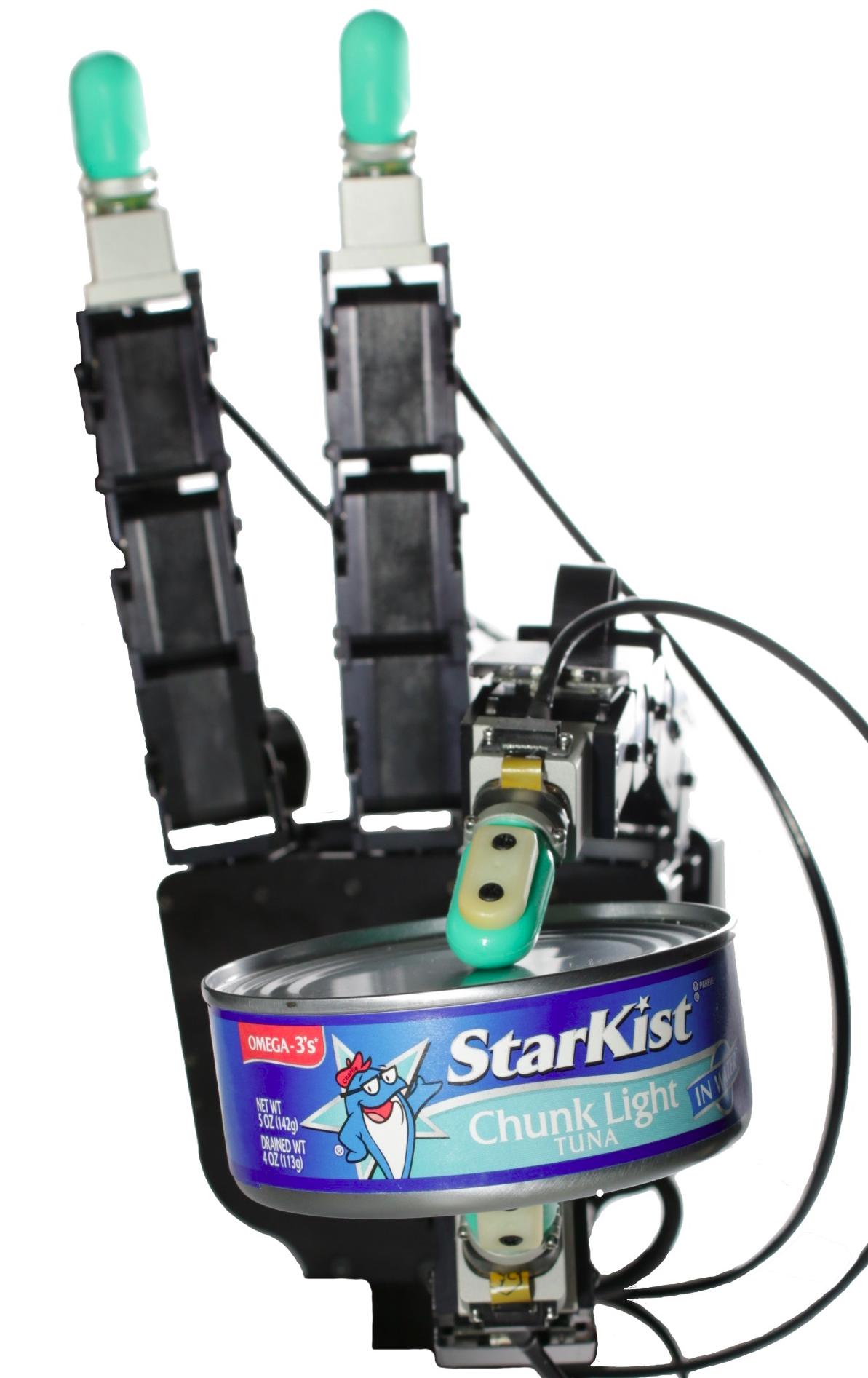}}\hspace{0.5cm}
    \subfloat[]{\includegraphics[width=0.25\linewidth]{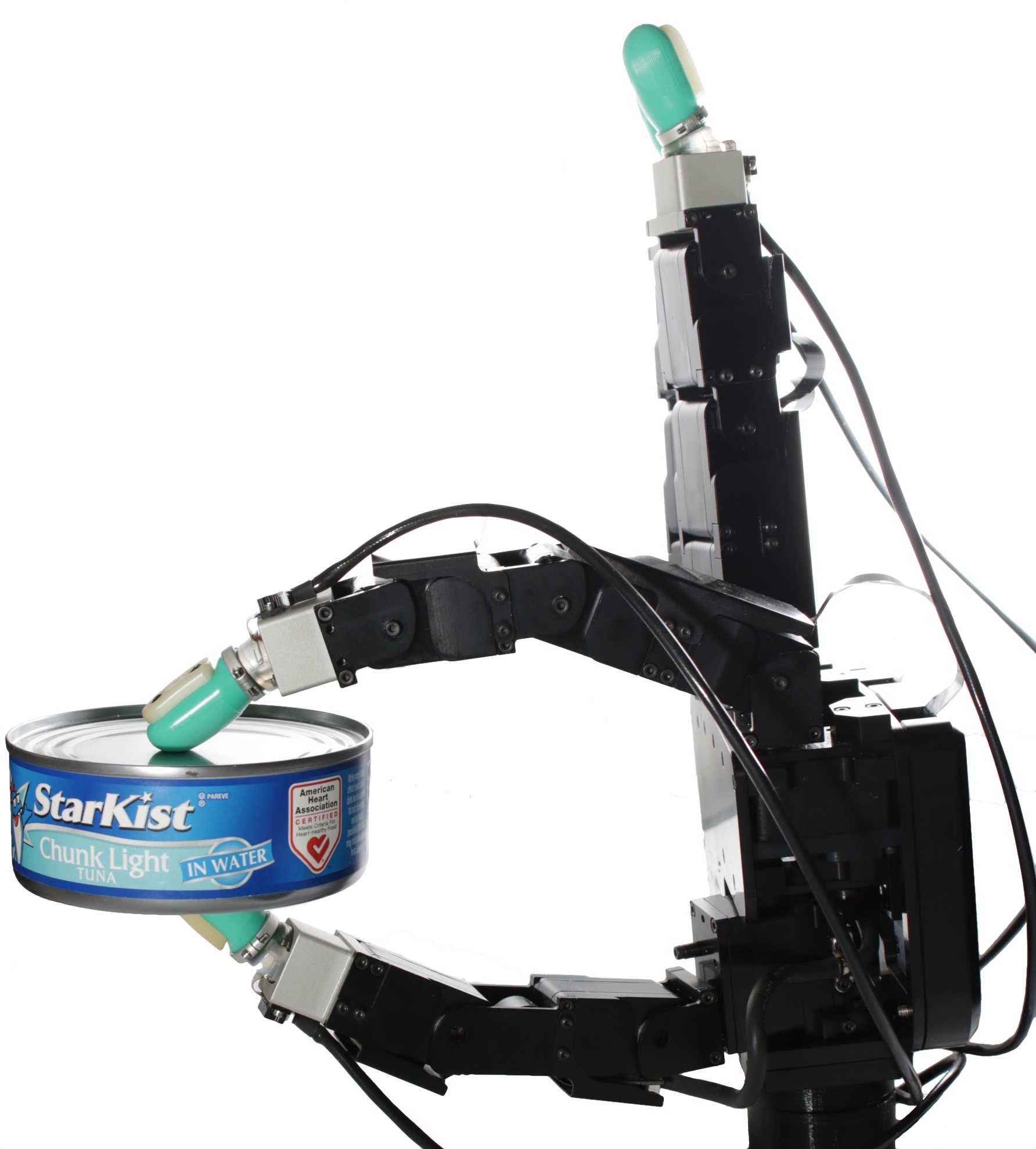}}\hspace{2cm}
    \subfloat[]{\includegraphics[width=0.2\linewidth]{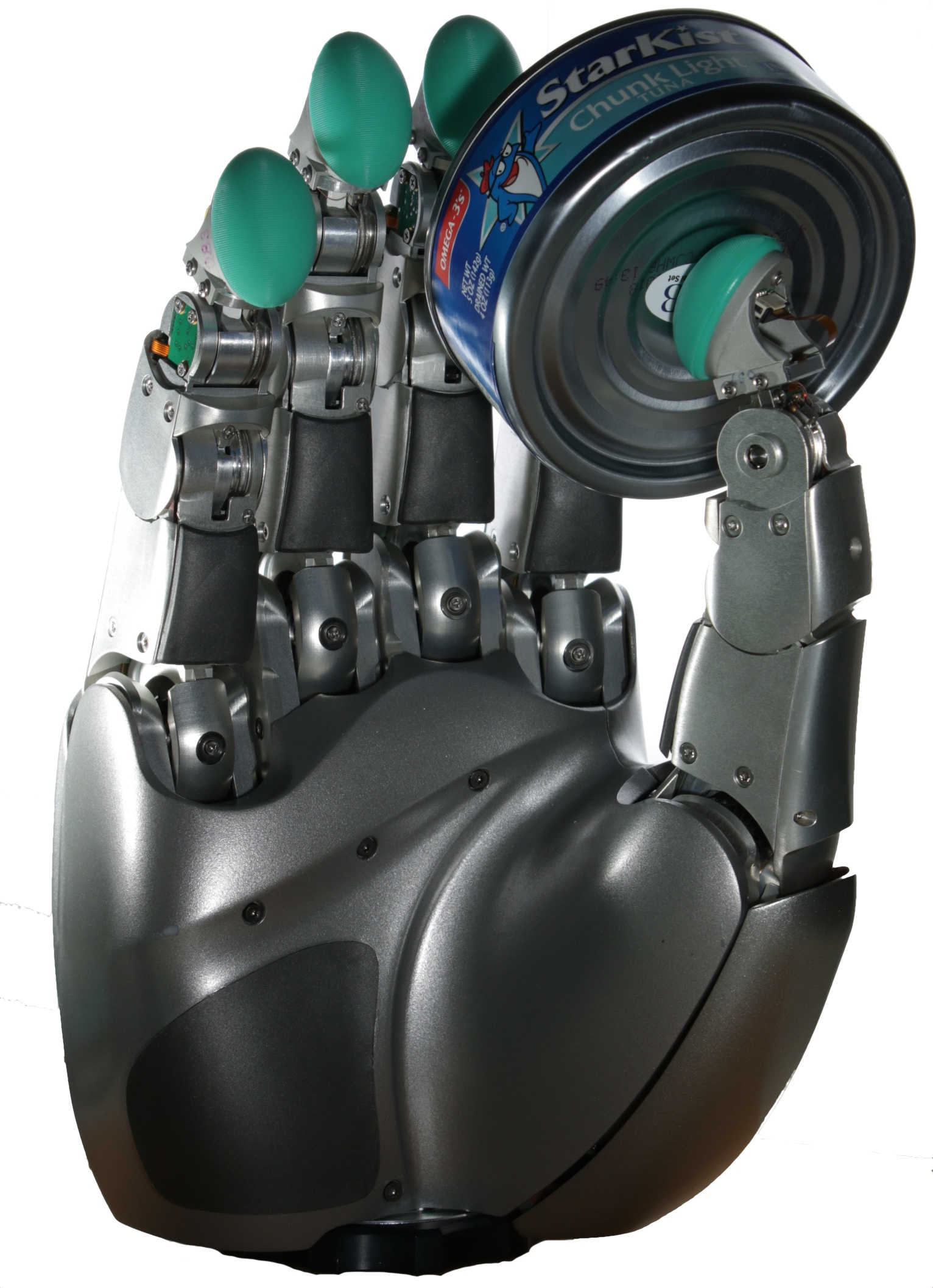}}\hspace{0.5cm}
    \subfloat[]{\includegraphics[width=0.2\linewidth]{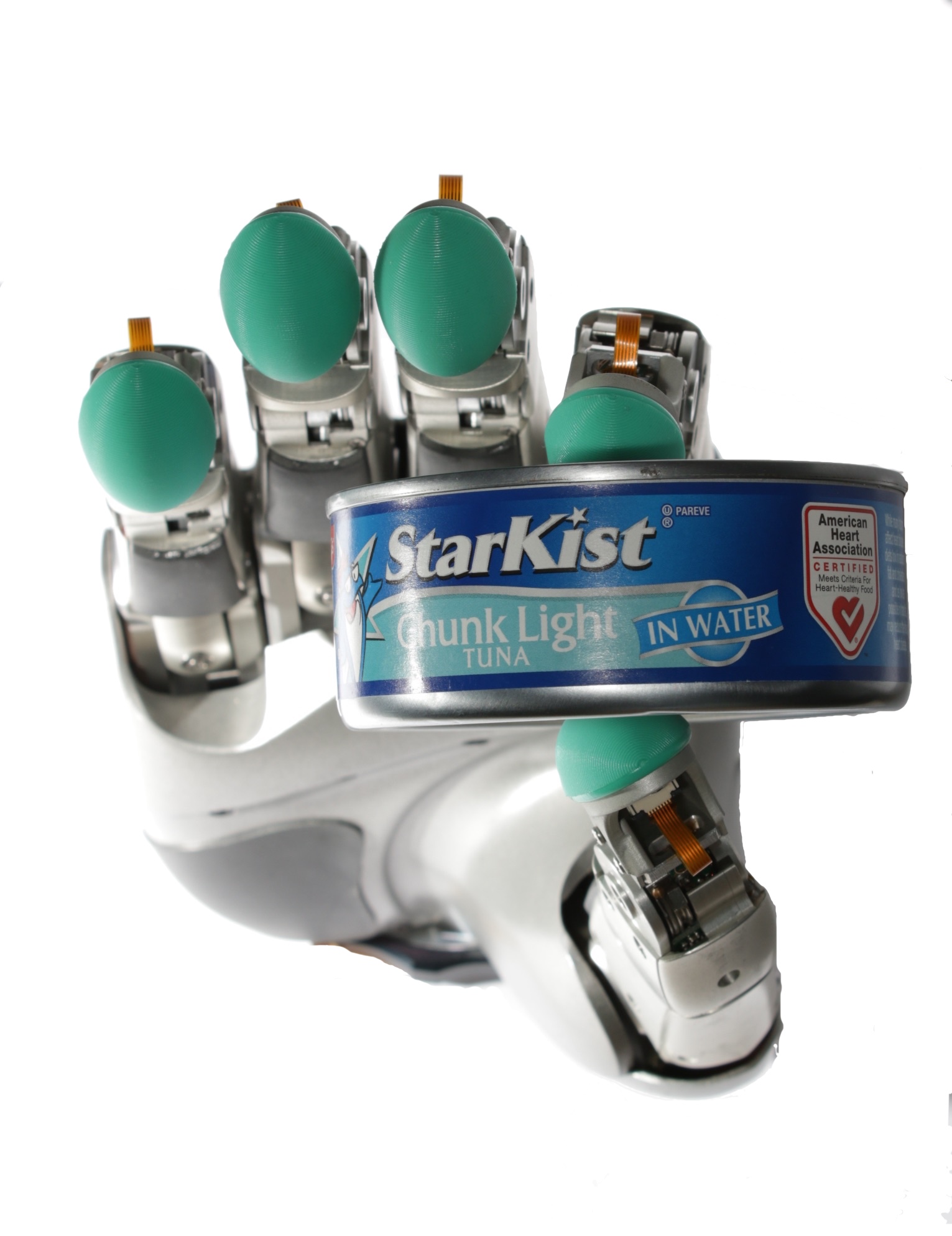}}\\
	(i) Two Finger Control\\
    \subfloat[]{\includegraphics[width=0.15\linewidth]{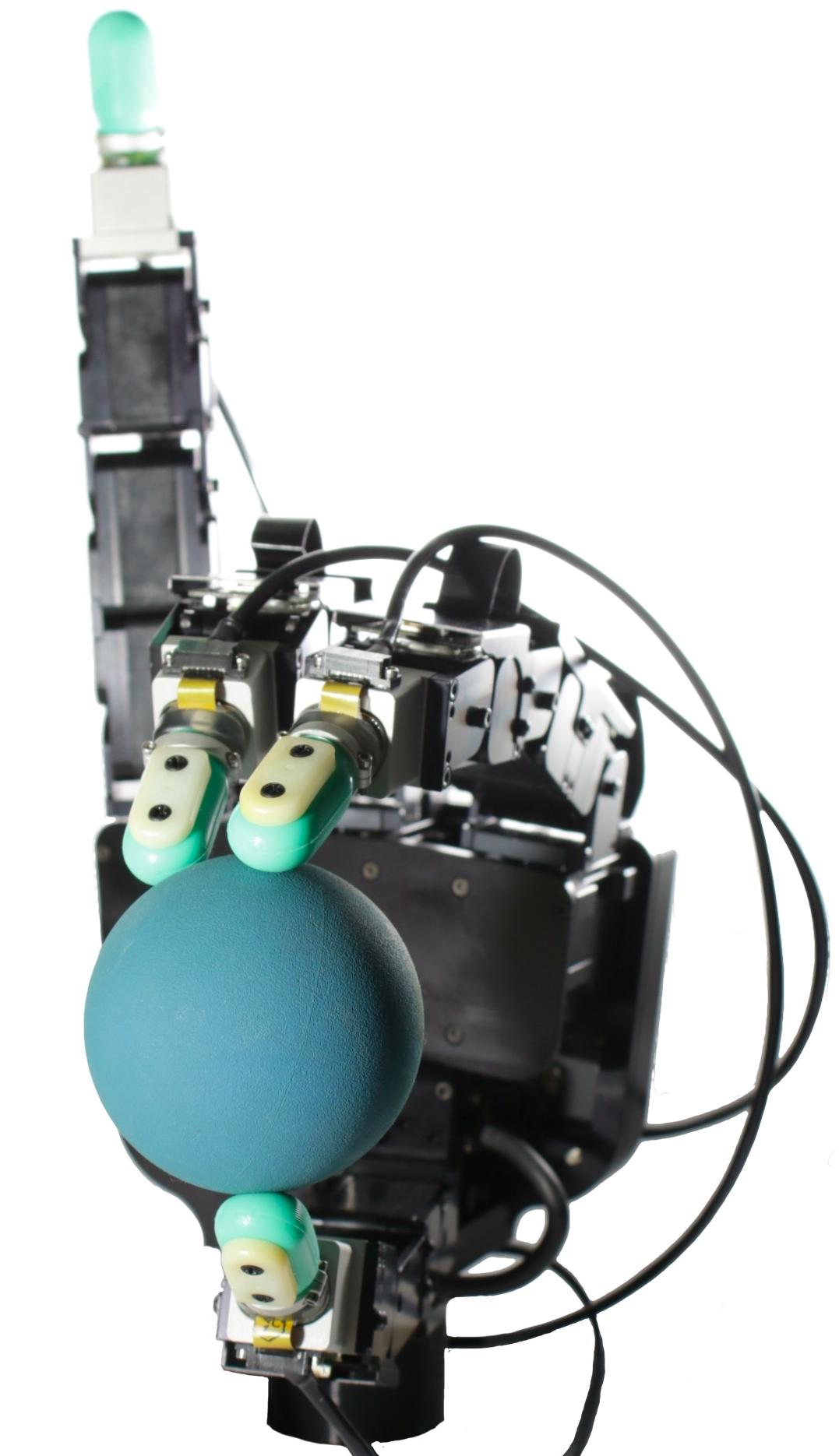}}\hspace{0.5cm}
    \subfloat[]{\includegraphics[width=0.25\linewidth]{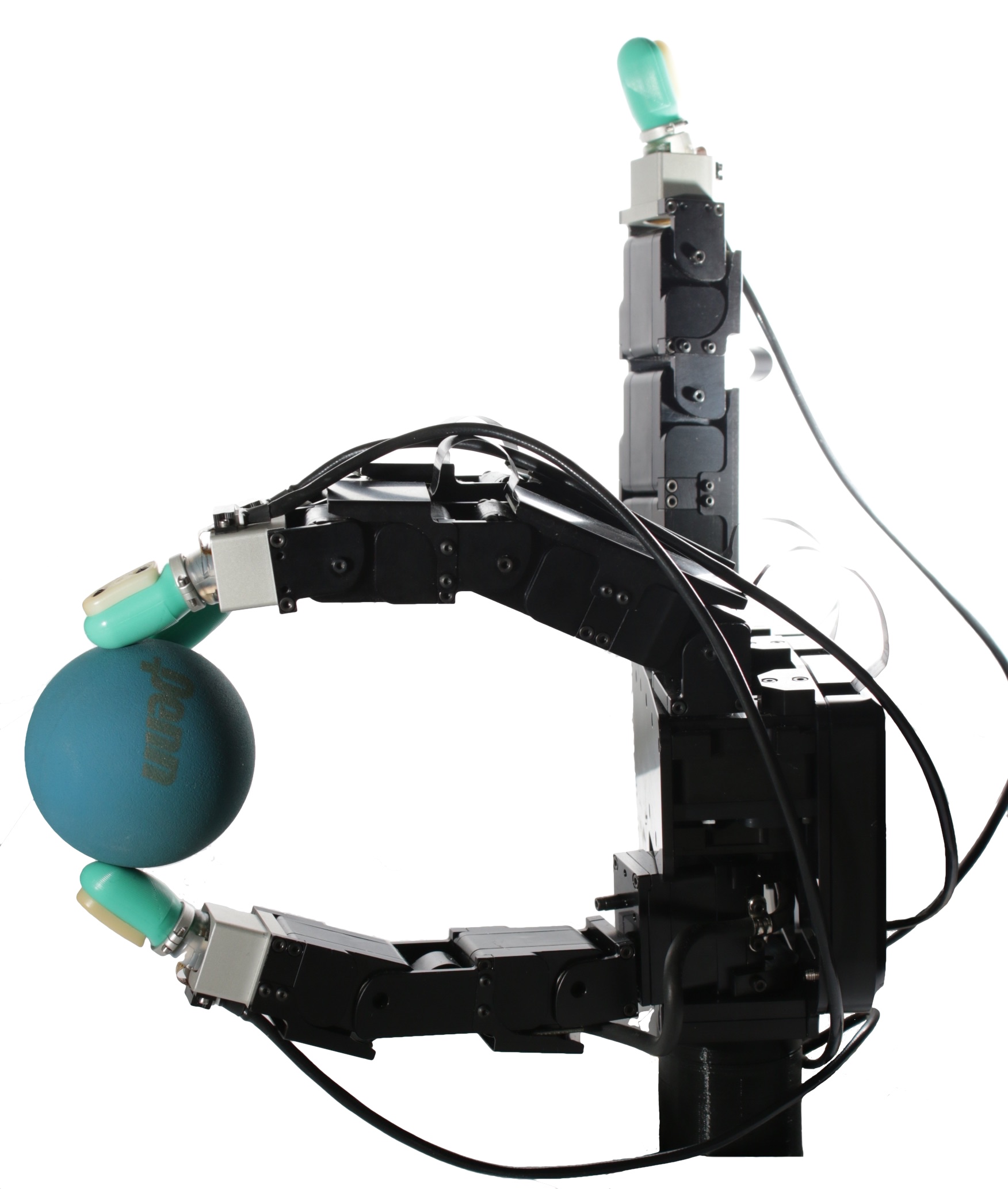}}\hspace{2cm}    
    \subfloat[]{\includegraphics[width=0.2\linewidth]{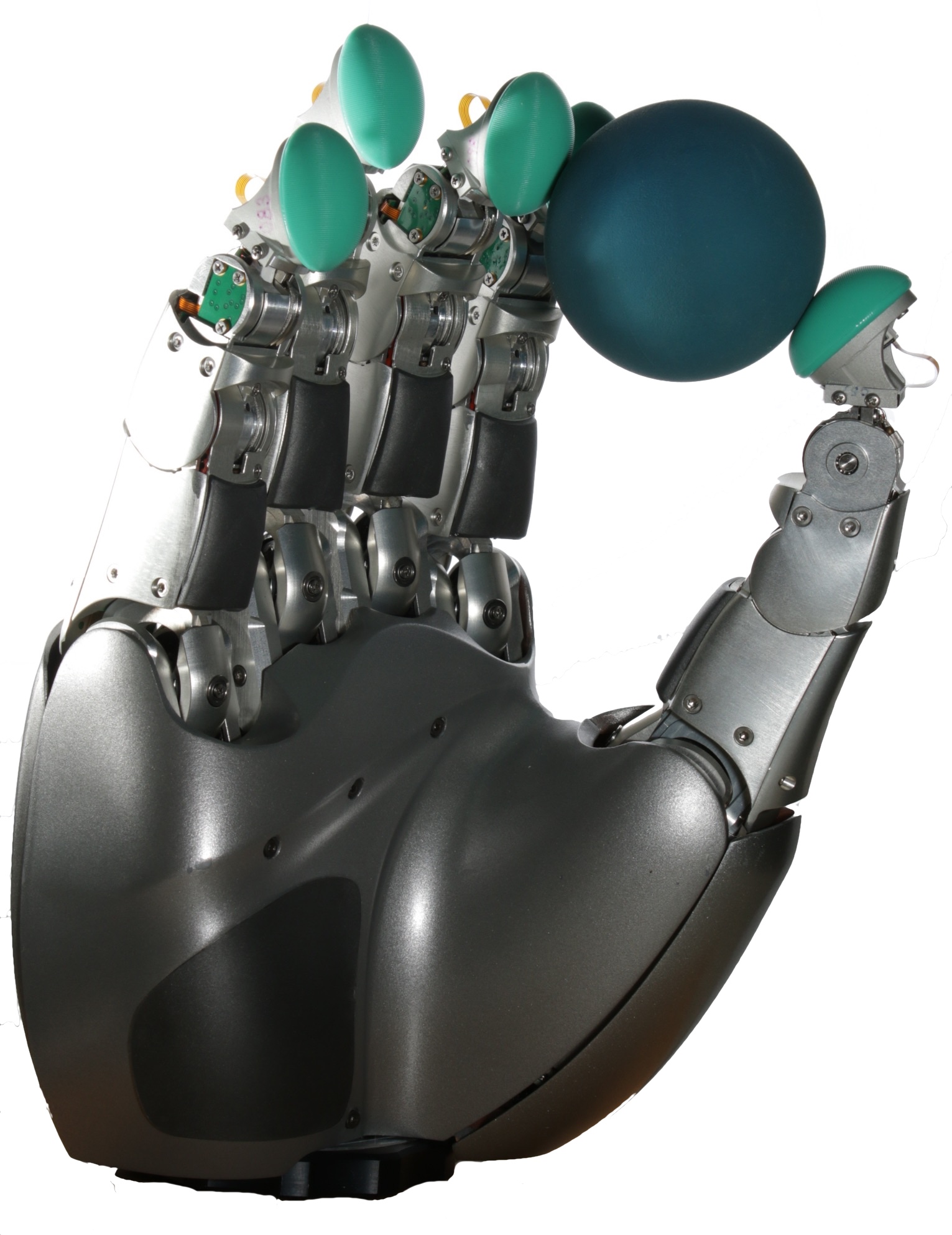}}\hspace{0.5cm}
    \subfloat[]{\includegraphics[width=0.2\linewidth]{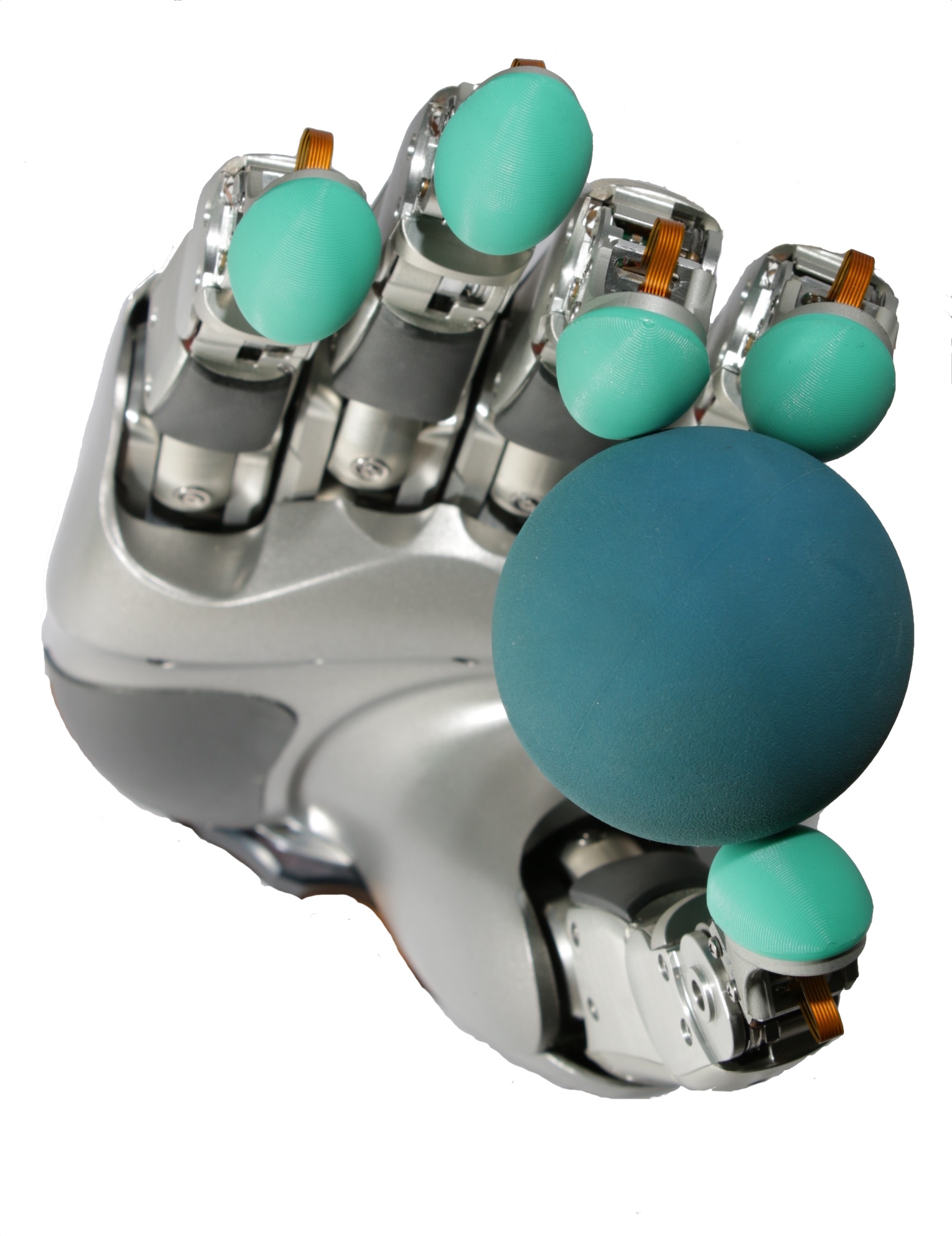}}\\
	(ii) Three Finger Control\\
    \hspace{-0.75cm}\subfloat[]{\includegraphics[width=0.2\linewidth]{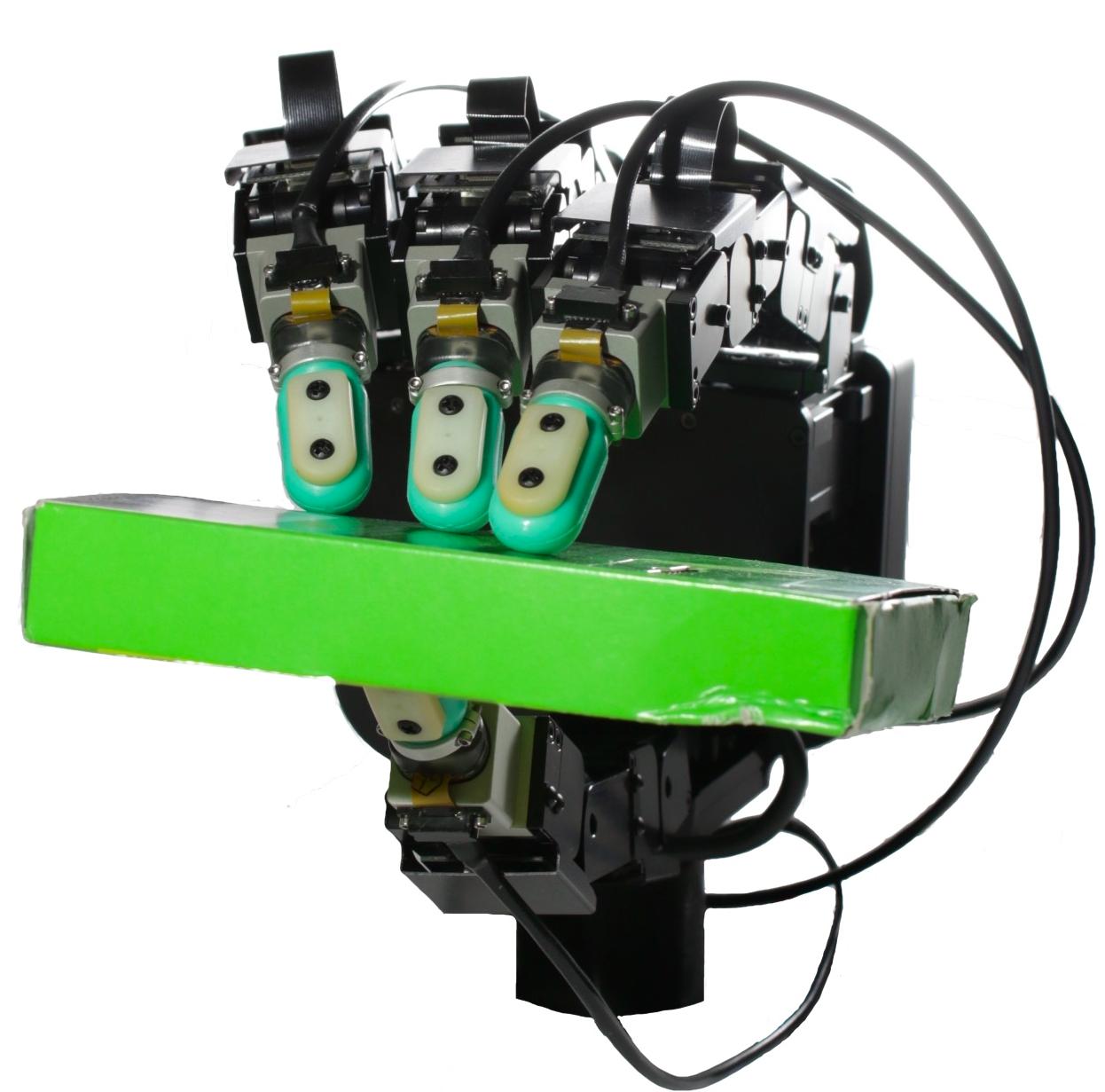}}\hspace{0.5cm}
    \subfloat[]{\includegraphics[width=0.25\linewidth]{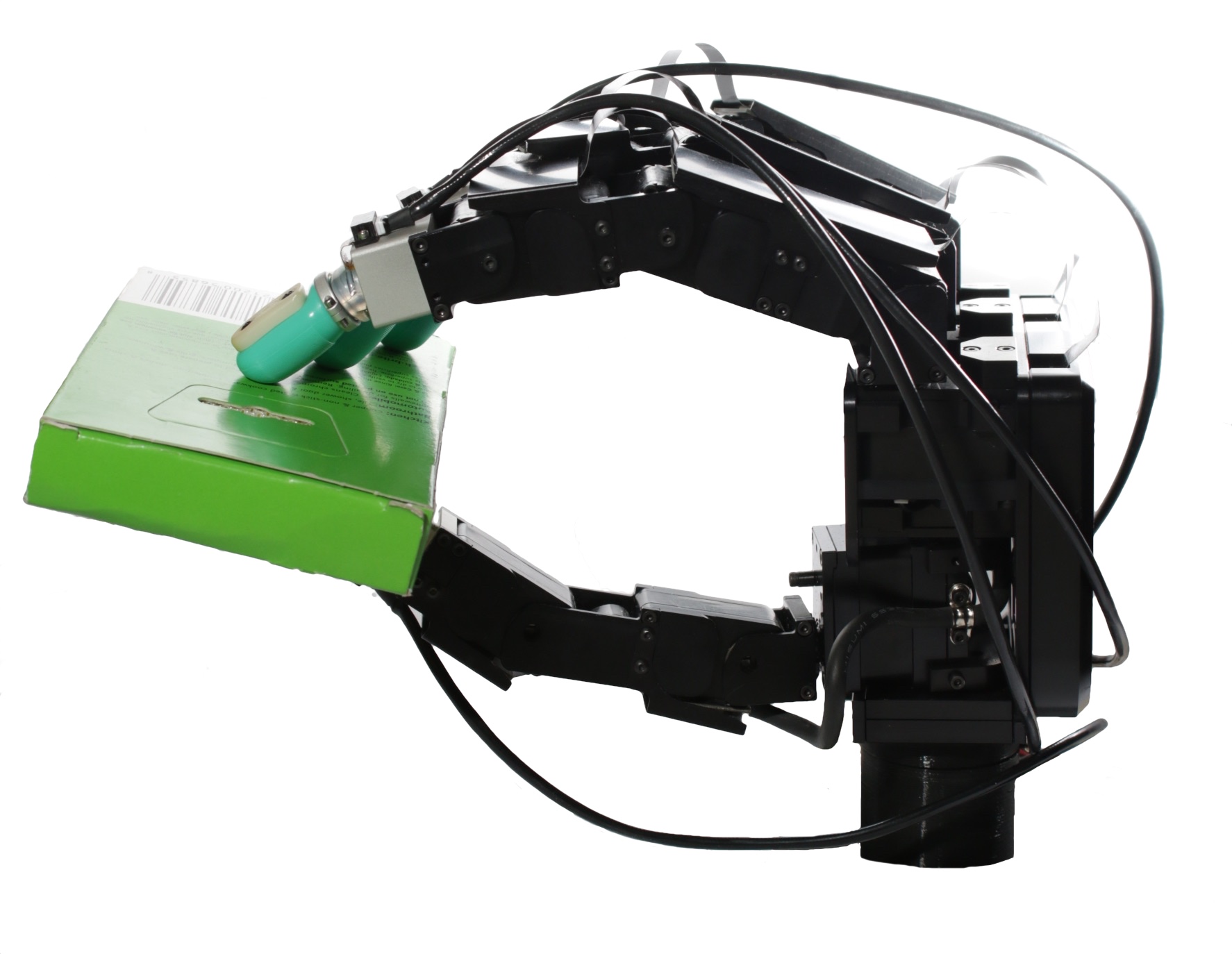}}\hspace{2cm}
    \subfloat[]{\includegraphics[width=0.2\linewidth]{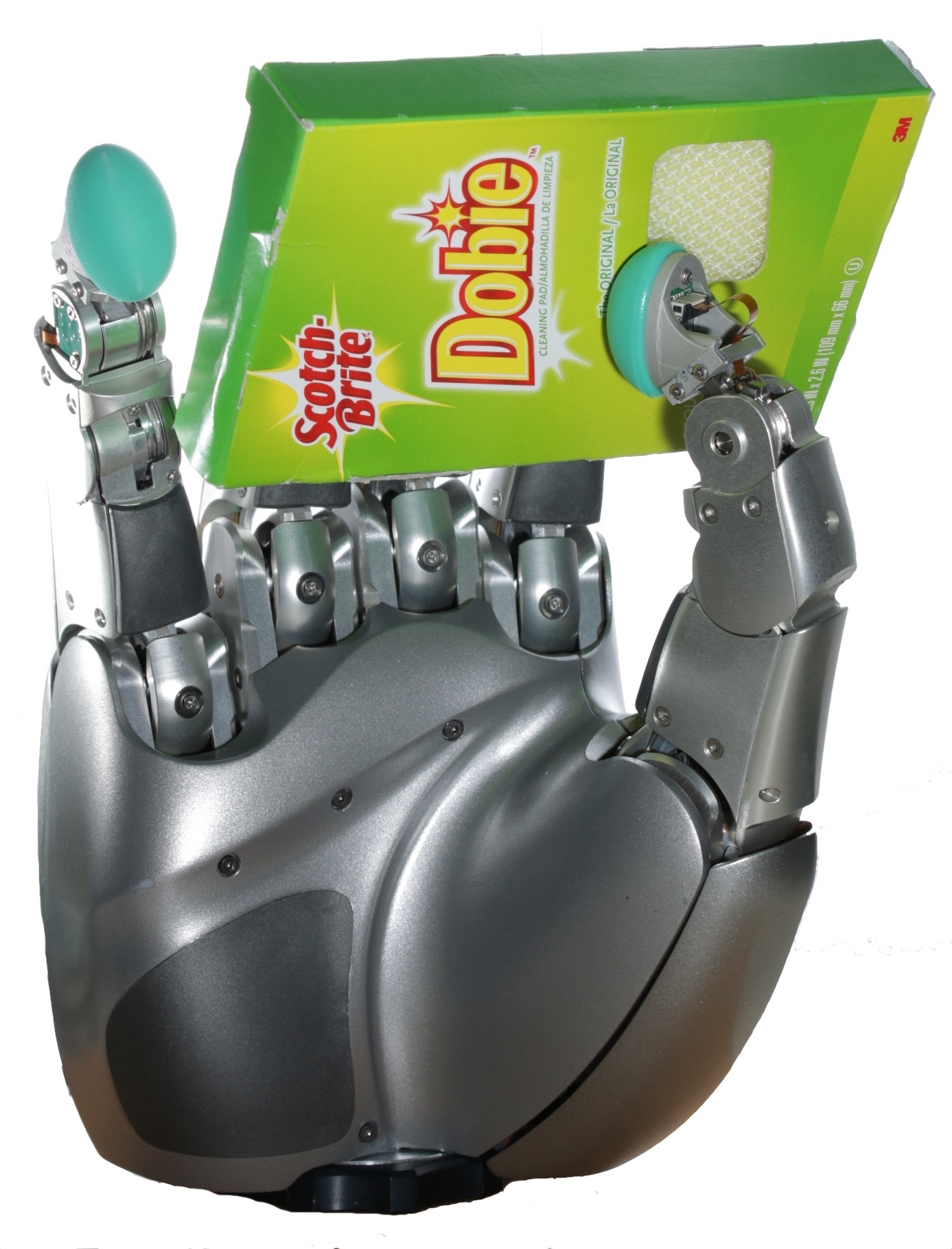}}\hspace{0.5cm}
    \subfloat[]{\includegraphics[width=0.2\linewidth]{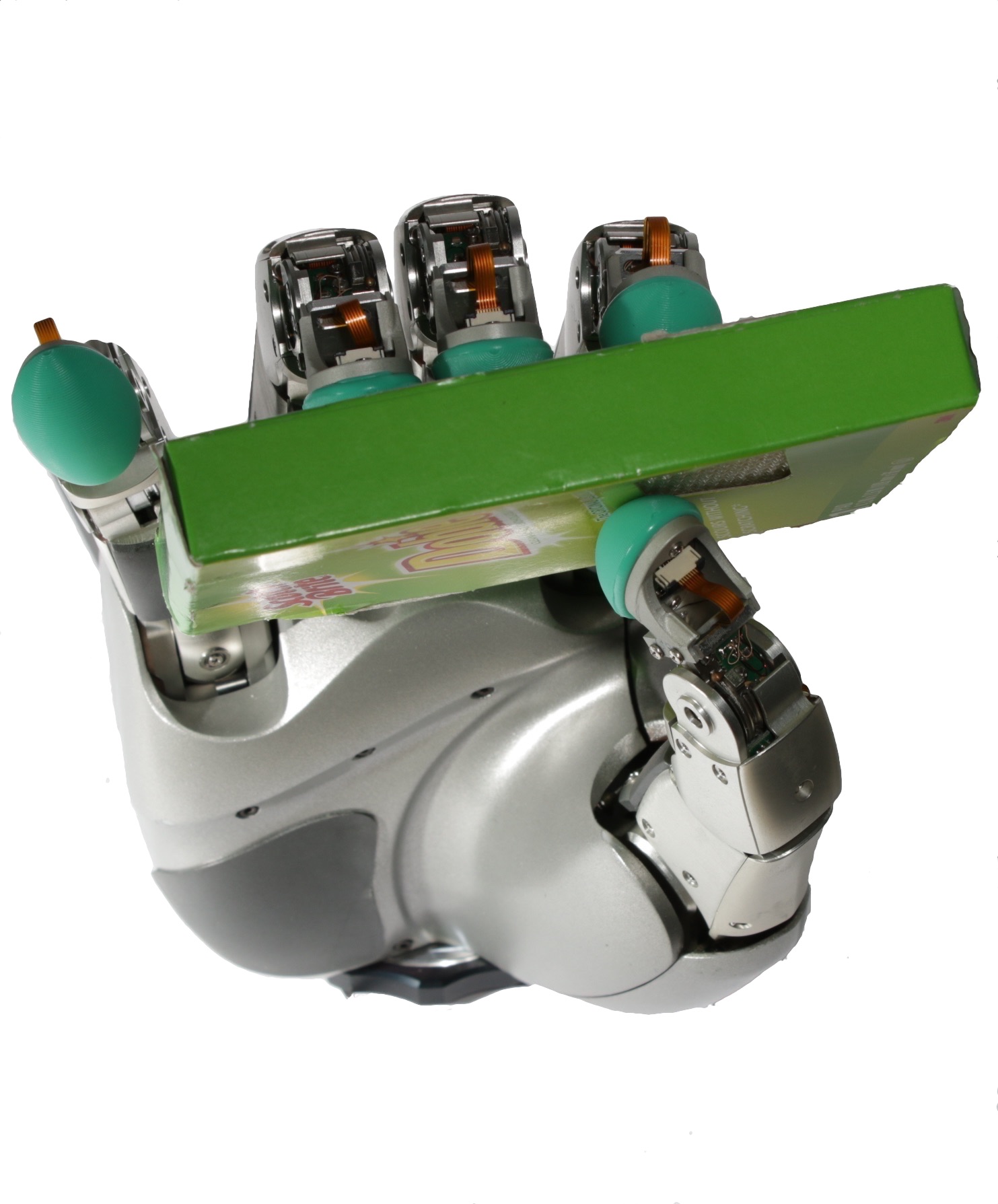}}\\
    (iii) Four Finger Control\\
    \subfloat[]{\includegraphics[width=0.2\linewidth]{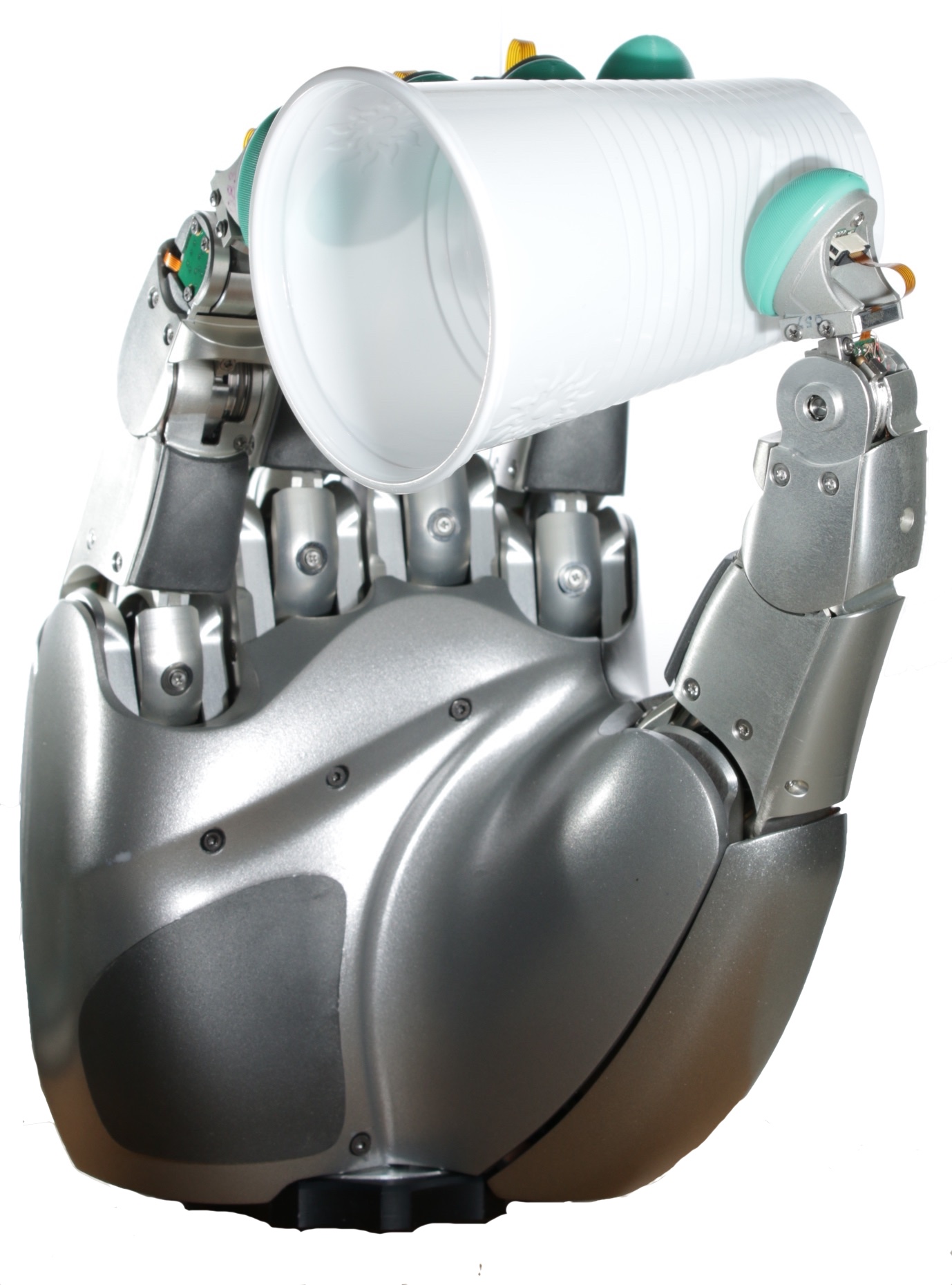}}\hspace{0.5cm}
    \subfloat[]{\includegraphics[width=0.2\linewidth]{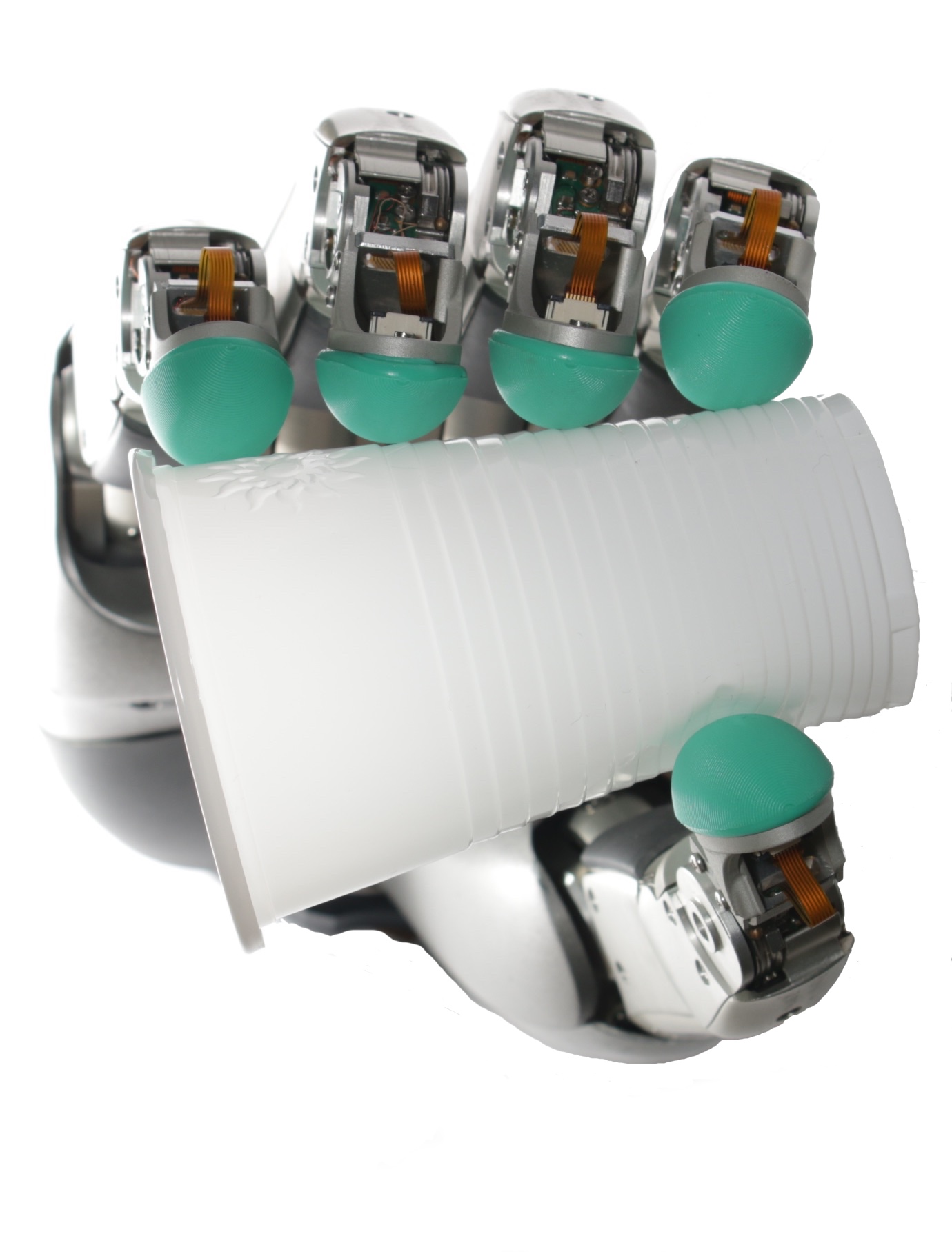}}\\
    (iv) Five Finger Control\\
    \caption{\label{fig:grip:stabilization:all:fingers}Stabilization of several objects using different grasp configurations with varying number of fingers. Both hands are controlled using independent controllers for each finger. In (i), a tuna can is stabilized using two fingers of the (a,b) Allegro and (c,d) Wessling Hands. In (ii), a tripod grasp is used to stabilize a ball using the (e,f) Allegro and (g,h) Wessling Hands. In (iii), four fingers are used to stabilize a cardboard box with the (i,j) Allegro and (k,l) Wessling Hands. Finally in (iv) the full five fingers of the Wessling hand are used to stabilize an empty plastic glass (m,n). For all grasps, the stabilization controllers of each finger were only trained with data collected from the ball and the cardboard box.}
    
\end{figure*}

The described procedure relies on randomly selected velocities in task space for the object surface servoing.   Target pressures are selected from three possible values from the set $P* = [20, 50, 80]$, in sensor grounded pressure units.
Spanning the data across multiple pressures in conjunction with randomly selected velocity parameters and the distinct contact locations observed across the three fingers, serve for training slip classifiers that are not characterized only by applied pressure, contact location or fingertip velocities. Additionally, all sensor values concerning either pressure or finger deformation are grounded before training, preventing parametric differences in the sensors (for example nominal fluid pressure) from correlating to slip.

\begin{figure}[b]
    \centering
    \includegraphics[width=\linewidth]{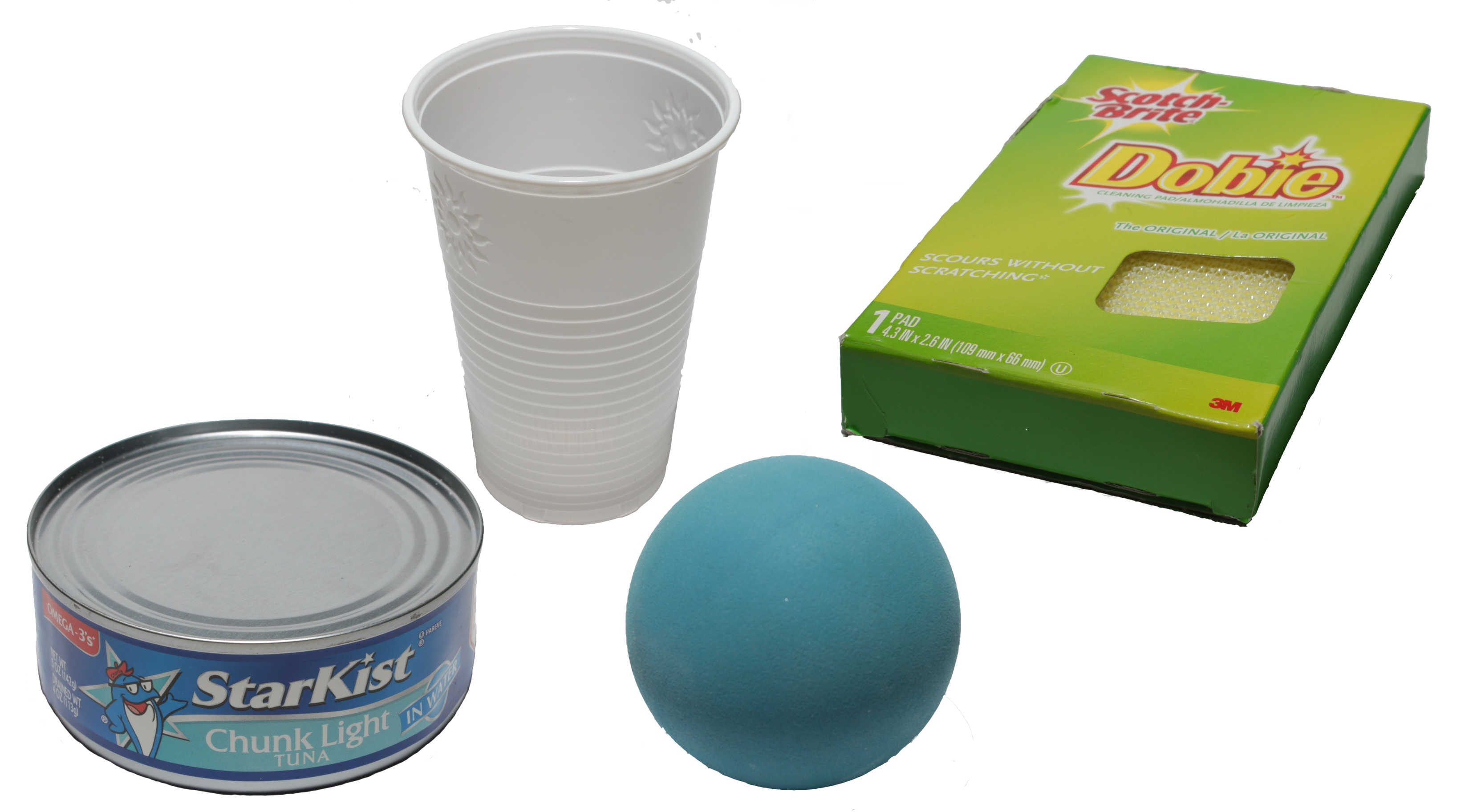}
    \caption{\label{fig:object:set}The four objects used in our experiments. While the ball and the box are used for training and testing, the plastic cup and the tuna can are only used for testing.}
\end{figure}

Four objects are used in our experiments and are depicted in Fig.~\ref{fig:object:set}. Note that from these for objects, only two of them (ball and cardboard box) are used for training the classifiers with the two remaining object only being used for testing.

Three trials are done for each value of $P*$ on the two separate objects for a total of 18 trials. Considering that three fingers are used per trial, the resulting data set correspond to 54 single finger trials. 

Labeling the data is done automatically by using the finger's end-effector location and the pressure values observed on each finger. Since the object is fixed, if the finger is in contact (observed pressure above a certain threshold $T_\textrm{Contact}$) and the finger is moving then we assume slip is occurring. A  minimum deviation $\Delta\mathbf{x}$ in finger position is enforced so that the skin deformation is taken into account.

\subsection{Grip Stabilization Evaluation} 
\label{sec:grip-stabilization-evaluation}
In the following section, results all possible stabilization configurations will be shown. The configurations include single finger human-robot stabilization and all possible multi-finger grip (two, three, four and five finger grips). For each configuration we will show the finger pressure behavior of each finger during the stabilization procedure.

External disturbances are applied on the object by a human subject in order to test grip robustness.

\subsubsection{One Finger Control}

The single finger object stabilization work that serves as base for this work was performed on a PA10 robot arm in~\cite{veigastabilizing}. Stabilization of an object jointly with a human can be seen in Fig.~\ref{fig:single:finger:stabilization}. A extensive study of the slip predictors performance and of the stabilization performance was performed.

\subsubsection{Two, Three and Four Finger Control}

For grip configurations involving two to four fingers, we attempted to stabilzise each of the objects while only having the slip predictors trainned with a subset of the objects. The predictors were trained with two of the objects (the ball and the box), having the remaining objects (the tuna can and the plastic cup) as unknown objects. These unknown objects were chosen based on their distinc properties with respect to the objects in the training set. The tuna can is the heaviest of the objects while the plastic cup is the most deformable.

All objects were successfully stabilized in hand with each of the grip configurations. Some examples of these grasps are shown in Fig.~\ref{fig:grip:stabilization:all:fingers}. 

%
%
%
%

\subsubsection{Five Finger Control}

For five fingered object stabilization, only the Wessling Hand can be used as the Allegro hand only possesses 4 digits. For this grip configuration we wish to particularly highlight the stabilization of the plastic cup. The cup is highly deformable and, due to the kinematics of the Wessling hand, there is very few oposition between the thumb, the ring and little fingers, increasing the difficulty of the five fingered grip. This lack of finger oposition, requires the little and ring fingers to apply a mimimum amount of force, otherwise causing the object to shift on the remaining fingers, rendering the grip unstable. Even under these constraints, the controllers are able to successfully stabilize the plastic cup, using the five fingers in the Wessling hand. The stabilization is not only successfull but is also performed without deforming the object's surface. The resulting five fingered grip is depicted in Fig.~\ref{fig:grip:stabilization:all:fingers}.

%% file: DiscussionConclusion.tex
\section{Conclusion and Discussion}
\label{sec:conclusion}

In this section we will provide an overview of the contributions of this paper, Sec~\ref{sec:summary-of-contributions}, describe the current short comings of the approach, Sec~\ref{sec:current-shortcomings} and outlining possible future directions for our work, Sec~\ref{sec:future-work}.

\subsection{Summary of the Contribution}
\label{sec:summary-of-contributions}

In this paper we presented a modular control approach capable of stabilizing objects in-hand using any number of fingers. Each finger was controlled independently, having the synchronization between fingers emerge from the tactile feedback of each finger controller. 
The tactile feedback allows for disturbances caused by poor contact distribution on the fingertip surface or introduced by other fingers action on the object to detected. Each finger will then automatically compensate for perceived changes that might compromise object stability, independently of these changes being cause by external sources, poor finger configurations or other fingers. 


Our modular control approach was shown to be generalizable across multiple objects, even objects that are fairly different to the objects in the training set. It is also effective across different platforms, has shown by the experiments performed on the two robot hands that have relevant differences in their kinematic chains. 

\subsection{Current Shortcomings}
\label{sec:current-shortcomings}

In this work, we wished to assess if our independence assumption was valid and how much it is possible to do with designed controllers enriched with the tactile signals. Using the low dimensional slip signals that where acquired in previous work, allowed for the design of the controller used in this paper. As the full tactile state is much richer than the slip signals, we are possibly discarding relevant information.  

Additionally, in this work we assumed that the tested objects were provided to the hand in configurations where the stabilization of said object would be possible, not requiring finger gating/re-positioning.

\subsection{Future Work}
\label{sec:future-work}
Partitioning the hand into a set of independent fingers allows the manipulation problem to be viewed as a distributed problem where each finger solves the problem locally, with simpler models than when considering a complete model for the full hand. This partitioning allows the models to be learned efficiently using data driven approaches that consider the the full sensor state, as the dimensionality of the problem is distributed across fingers. Our future work will focus on exploring the high dimensionality of the feedback signals and learning stabilization controllers using reinforcement learning approaches in these high dimensional spaces.

As each finger has only the goal of stabilizing the object while remaining in contact with it, manipulating an object would simply require one of the fingers to introduce the desired perturbation to the object, while the remaining fingers keep it stable. Exploring the assumption that small manipulations can be achieved by this kind of master slave paradigms is another interesting future work direction.

Finally, for complex manipulations, we believe that independently controlling the fingers will be necessary but not sufficient to achieve robust performance. Nonetheless, using the independent control as the base level in a hierarchical control framework might enable higher level control policies to perform these manipulations, effectively creating a robust control hierarchy, where the task complexity is distributed across the several levels of the hierarchy. Building such a hierarchy is also viewed as potentially interesting future work.

%% file: Acknowledgements.tex
\section{Acknowledgments}

The research leading to these results has received funding from the European Community’s Seventh Framework Programme (FP7/2007–2013) under grant agreement 610967 (TACMAN).